\documentclass[runningheads]{llncs}
\usepackage[T1]{fontenc}
\usepackage{nopageno}
\usepackage[utf8]{inputenc} 
\usepackage[T1]{fontenc}    
\usepackage{hyperref}       
\usepackage{url}            
\usepackage{booktabs}       
\usepackage{amsfonts}       
\usepackage{nicefrac}       
\usepackage{microtype}      
\usepackage{lipsum}
\usepackage{tabularx}
\usepackage{amsmath}
\usepackage{subcaption}
\usepackage{fancyhdr}       
\usepackage{xspace}
\usepackage[pdftex]{graphicx}
\usepackage{lmodern}
\usepackage{subcaption}
\newcommand{\latinphrase}[1]{\textit{#1}}
\newcommand{\etal}{\latinphrase{et~al.}\xspace}

\title{xLSTM-FER: Enhancing Student Expression Recognition with Extended Vision Long Short-Term Memory Network}

\begin{document}
\author{Qionghao Huang\inst{1} \and
Jili Chen\inst{2}}


\institute{Zhejiang Key Laboratory of Intelligent Education Technology and Application,
Zhejiang Normal University, Zhejiang Province, Jinhua, China
\email{qhhuang@zjnu.edu.cn}\\ \and
Zhejiang Key Laboratory of Intelligent Education Technology and Application,
Zhejiang Normal University, Zhejiang Province, Jinhua, China\\
\email{irelia@zjnu.edu.cn}
}
%
\index{Huang, Qionghao}
\index{Chen, Jili}

\maketitle

\begin{abstract}
Student expression recognition has become an essential tool for assessing learning experiences and emotional states. This paper introduces xLSTM-FER, a novel architecture derived from the Extended Long Short-Term Memory (xLSTM), designed to enhance the accuracy and efficiency of expression recognition through advanced sequence processing capabilities for student facial expression recognition. xLSTM-FER processes input images by segmenting them into a series of patches and leveraging a stack of xLSTM blocks to handle these patches. xLSTM-FER can capture subtle changes in real-world students' facial expressions and improve recognition accuracy by learning spatial-temporal relationships within the sequence. Experiments on CK+, RAF-DF, and FERplus demonstrate the potential of xLSTM-FER in expression recognition tasks, showing better performance compared to state-of-the-art methods on standard datasets. The linear computational and memory complexity of xLSTM-FER make it particularly suitable for handling high-resolution images. Moreover, the design of xLSTM-FER allows for efficient processing of non-sequential inputs such as images without additional computation.
\end{abstract}

\section{Introduction}
Student facial expression recognition is a burgeoning field with significant implications for educational technology. By analyzing students' facial cues, educators can gain insights into their emotional states, engagement levels~\cite{tongucc2020automatic},  cognitive load~\cite{jagadeesh2022facial}, and academic performance~\cite{huang2024enhancing,huang2024improving} during learning activities~\cite{huang2019adaptive}. The current student face recognition systems primarily include those based on traditional CNN-based and Vision Transformer~\cite{dosovitskiy2020image} (ViT)-based approaches.
The lightweight and efficient characteristics of CNNs have attracted the attention of early education researchers, leading to the development of a series of face recognition systems and teaching environments based on CNNs~\cite{ozdemir2019real}. The ViT has replaced CNN as a more robust backbone network for student facial expression recognition. The Vision Transformer, leveraging self-attention for global image modeling, has surpassed the performance of CNNs in both teaching feedback systems~\cite{wang2024teaching} and the assessment of learning outcomes~\cite{huang2021facial,wu2023fer}.

However, for CNNs, the main limitation is their lack of global receptive fields and dynamic weighting capabilities, which can restrict their ability to capture long-range dependencies and integrate information from the entire image~\cite{jiang2023face2nodes}. Besides, this advantage of ViTs comes at the cost of quadratic complexity in terms of image sizes, which leads to a significant computational overhead when dealing with dense prediction tasks such as object detection and semantic segmentation~\cite{zhou2023emotion}.

To address the aforementioned issues, we propose the xLSTM-FER. xLSTM-FER begins by segmenting the input image into a series of non-overlapping patches, converting the 2D image into a 1D token sequence with added learnable 2D positional encodings to retain spatial information. These sequences are then fed into an xLSTM encoder composed of stacked xLSTM blocks. The xLSTM blocks maintain a linear complexity while capturing long-range dependencies and spatial-temporal dynamics within the image sequence. Each xLSTM block employs a modified LSTM layer (mLSTM) that uses matrix values for memory retrieval, enhancing the model's capacity to discern subtle facial movements. To overcome the inherent difficulty of parallel processing in LSTM, the mLSTM utilizes a memory matrix to enhance parallel capabilities. By integrating different path traversals, the model achieves a comprehensive image representation. The summary of our contributions is as follows:

\begin{itemize}
    \item We propose xLSTM-FER, which segments input images into a series of patches and processes them through a stack of xLSTM blocks, allowing the model to capture subtle facial expression changes and improve recognition accuracy by learning the spatial-temporal dynamics within the sequence.
    \item The xLSTM-FER has the capabilities of parallelization and scalability through the memory matrix calculation. With its linear computational and memory complexity,  which is essential for capturing clear and detailed student expressions and making xLSTM-FER a more practical solution for real-world applications. 
    \item The extensive empirical evaluations of the xLSTM-FER model on multiple standard datasets demonstrate its superior performance in facial expression recognition tasks, including a perfect score on the CK+~\cite{lucey2010extended} dataset, and shows substantial improvements over previous state-of-the-art methods on both RAF-DB~\cite{li2017reliable} and FERplus~\cite{barsoum2016training} datasets.
\end{itemize}

\section{Related Work}
\label{sec:related work}

\subsection{Student Facial Expression Recognition in Learning Environment}
Early work utilizes Convolutional Neural Networks (CNNs) as the backbone for facial expression recognition tasks. Mohamad~\etal~\cite{mohamad2020automatic} use a VGG-B network to calculate the level of student engagement in MOOCs based on their facial expressions. Lasri~\etal~\cite{lasri2019facial} demonstrate a CNN-based automatic facial recognition system in educational settings can assist teachers in adjusting their teaching strategies and materials according to the emotional responses of students. Wang~\etal~\cite{wang2023facial} introduce a framework integrating an enhanced MobileViT~\cite{mehta2021mobilevit} model with an online platform for real-time student emotion analysis. To analyze student expressions and inform teaching strategies, Ling~\etal~\cite{ling2021facial} present a classroom-based FER system using YOLO and ViT. The computational demands of ViTs grow quadratically with the self-attention mechanism, which can be prohibitive for applications requiring high-resolution processing. To make facial recognition more efficient in educational scenarios, xLSTM-FER demonstrates linear computational and memory complexity, making it more suitable for training and practical deployment.

\subsection{Long Short-Term Memory Network}
LSTM~\cite{hochreiter1997long} is a type of recurrent neural network (RNN) architecture that is particularly good at learning order dependence in sequence prediction problems. Recently, Beck~\cite{beck2024xlstm} propose improvements to LSTM, including exponential gating and novel memory structures, to address the limitations of LSTM and enable it to scale to larger model sizes. Alkin~\etal~\cite{alkin2024vision} verify that xLSTM is also applicable as a visual backbone network.  Compared to CNNs, xLSTM has the characteristic of being scalable, and compared to Vision Transformers, it has a more linear complexity which makes it easier to deploy in practice. However, its application in student facial expression recognition remains unexplored. Therefore, we propose xLSTM-FER to explore the potential application of LSTM-based models and overcome the challenge in student expression recognition.

\section{Methodology}
\label{sec:method}

\subsection{Patch Embedding}
\begin{figure*}[!htb]
    \centering
    \begin{subfigure}{0.48\textwidth}
        \includegraphics[width=0.85\textwidth]{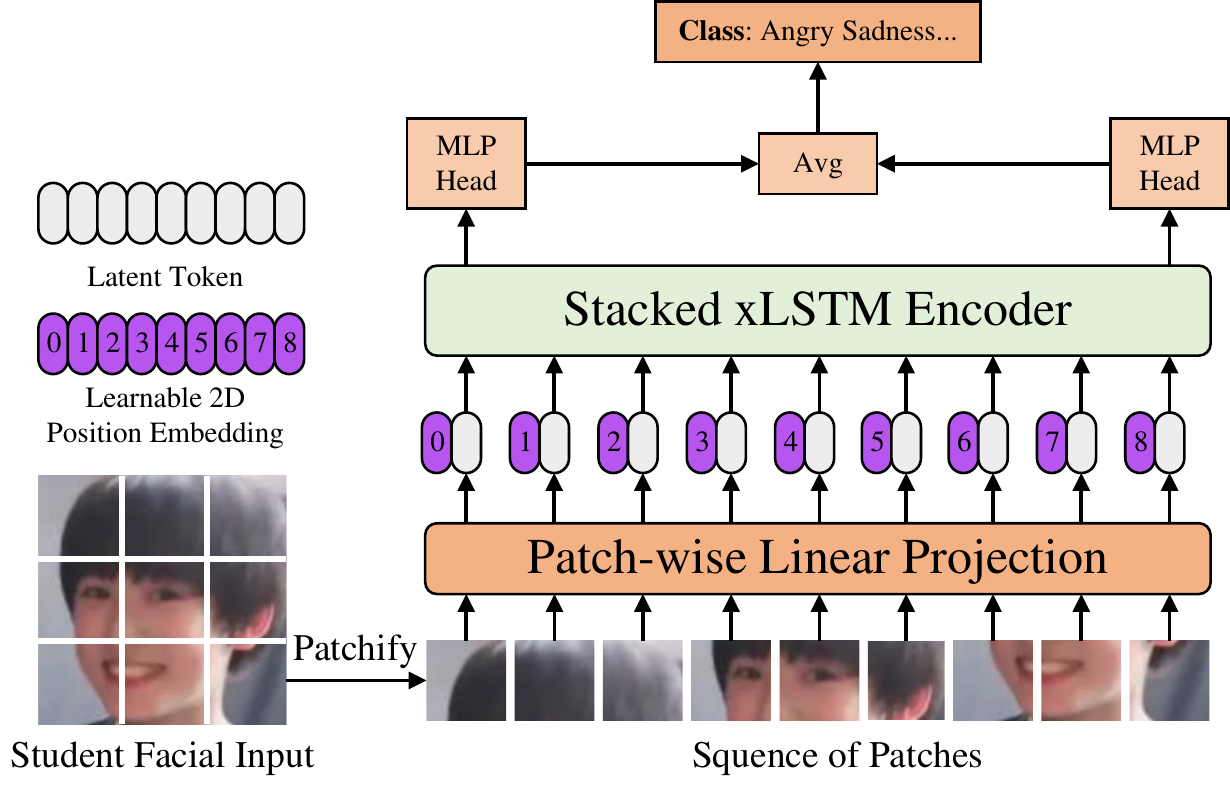} 
        \caption{The pipeline of our xLSTM-FER.}
        \label{fig:architecture}
    \end{subfigure}
    \hfill
    \begin{subfigure}{0.48\textwidth}
        \includegraphics[width=0.85\textwidth]{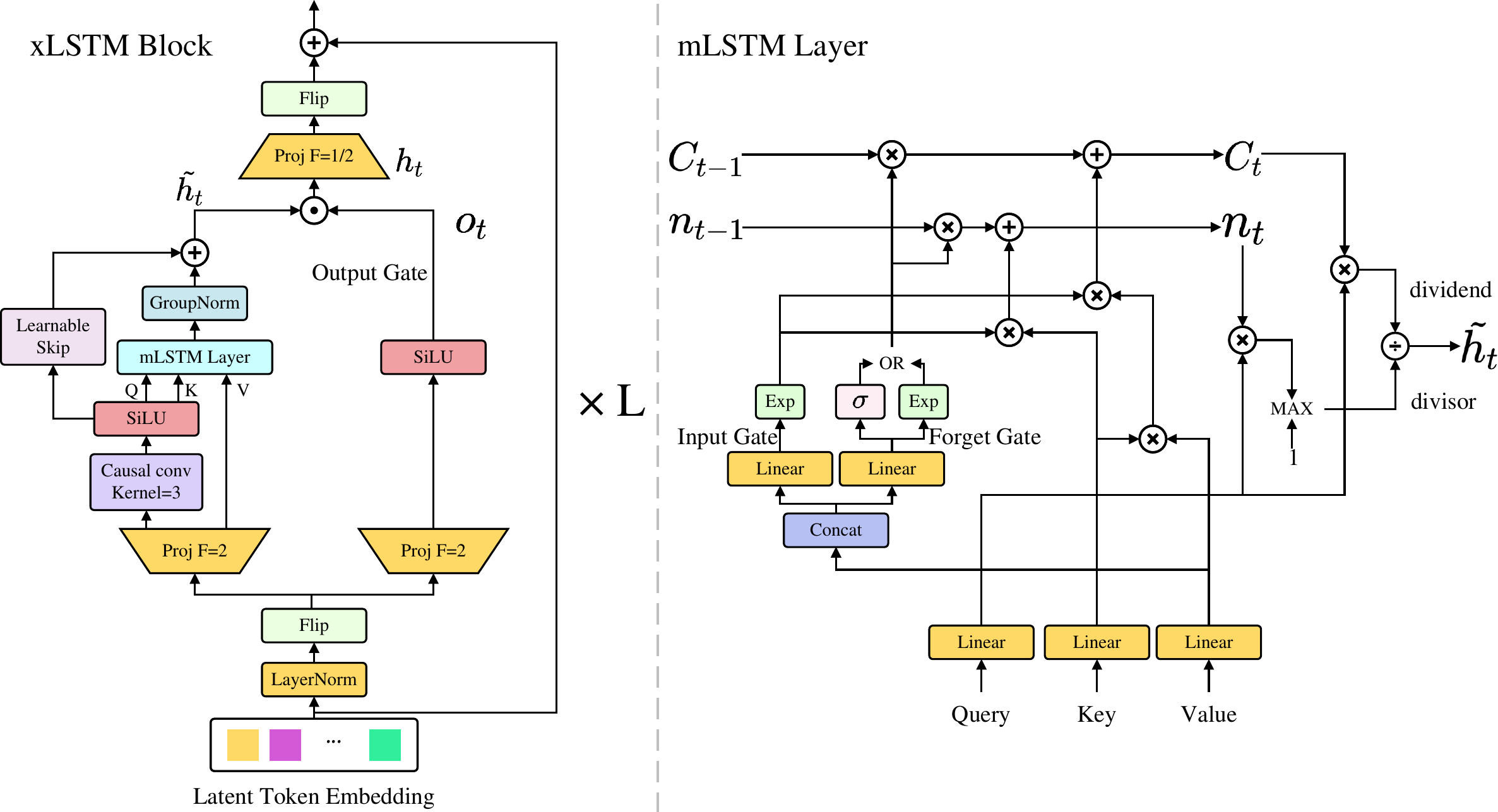} 
        \caption{Left: The model diagram of the xLSTM Block. Right: The model diagram of the xLSTM Layer.}
        \label{fig:block}
    \end{subfigure}
    \caption{Framework of our xLSTM-FER.}
\end{figure*}

The overall architecture of our network is shown in Fig.~\ref{fig:architecture}. We first perform patchification on the image. The input image $x\in \mathrm{R}^{H\times W\times C}$ is divided into a grid of non-overlapping patches. Each patch is a small square or rectangle of pixels with a width of $P$. Then, each patch is flattened into a sequence of pixel values $X_p\in \mathrm{R}^{N\times (P^2 \times C)}$, where $N=HW/(P^2)$. The flattened patch sequences are then linearly projected to a higher-dimensional space. To provide the model with information about the relative positions of the patches, we add learnable 2D positional embeddings to the patch sequences.

\subsection{xLSTM Encoder}
\textbf{xLSTM Block.} 
The xLSTM encoder is a structure composed of $L$-layer stacked xLSTM Blocks, as shown in Fig.~\ref{fig:block}. The xLSTM Block begins by layer-normalizing and then inverting the input. One branch doubles the channels (F=2) to construct the output gating, while the other branch uses a causal convolution layer (Kernel=3) to build the input for the mLSTM layer, which includes the query and key branch vectors for the linear attention mechanism~\cite{katharopoulos2020transformers}, with the value vector bypassing the causal convolution. The output of the mLSTM layer is passed through a group normalization layer~\cite{wu2018group} and is then summed with the output of the causal convolution via a weighted residual connection to obtain $\tilde{h_t}$, and $\tilde{h_t}$ is gated with the result of the output gate $o_t$ to obtain the output of the hidden state $h_t$. Finally, the channels are halved (F=1/2) and sum with the input embeddings of the block through a residual connection to obtain the entire xLSTM block's output. This high-capacity storage capability enables the model to distinguish between subtle differences in facial expressions,  crucial for identifying even the most nuanced emotions. This scalability is essential for creating robust systems capable of operating in diverse real-world environments.

\textbf{mLSTM Layer.} The mLSTM employs a FlashAttention mechanism, which is simulated using query, key, and value to guide the updates of both the cell state and the normalizer state, and subsequently outputs the results of the hidden layer as illustrated in Fig.~\ref{fig:block}. Specifically, the mLSTM layer first performs linear projections on the query, key, and value vectors:

\vspace{-10px}
\begin{align}
\begin{aligned}\label{eq:qkv proj}
    \operatorname{Query Mapping}\quad q_{t}\,=\,W_{q}\,x_{q}\,+\,b_{q},\\
    \operatorname{Scaled Key Mapping}\quad k_{t}\;=\;\frac{1}{\sqrt{d}}W_{k}\;x_{k}\;+\;b_{k},\\
    \operatorname{Value Mapping}\quad v_{t}\,=\,W_{v}\,x_{v}\,+\,b_{v},
\end{aligned}
\end{align}
where $x_q$, $x_k$, and $x_v$, represent the input query, key, and value vectors respectively, while $W_{q}$, $W_{k}$, and $W_{v}$ are the corresponding mapping matrices. $b_{q}$, $b_{k}$, and $b_v$ are the corresponding bias terms. By concatenating the mapped query, key, and value vectors, the input $x_t$ is obtained for the memory network to perform memory updates. The xLSTM uses an input gate and a forget gate to control the situation of memory updates and employs exponential gating and OR gating to facilitate the matrix memory calculation:

\vspace{-20px}

\begin{align}
\begin{aligned}\label{eq:IF gate}
\operatorname{Input\, Gate:}\quad i_t=\exp(\tilde{i}_{t}),\quad \tilde{i}_{t}=w_{i}^{\top}x_t+b_i,\\
\operatorname{Forget\, Gate:}\quad f_t=\exp(\tilde{f}_{t}) \operatorname{OR} \sigma({f}_{t}) ,\quad \tilde{f}_{t}=w_{f}^{\top}x_t+b_f,\\
\end{aligned}
\end{align}
where $w_{i}^{\top}$, $w_{f}^{\top}$, $b_i$, $b_f$ denote the weight vectors and bias terms corresponding to the input gate and forget gate, respectively. The $\sigma$ denotes the activation function, and $\exp(\cdot)$ signifies the exponential operation. The mLSTM expands the memory cell into a matrix. By integrating the update mechanism of LSTM with the information retrieval scheme from Transformers, mLSTM introduces an attention-integrated cell state and hidden state update scheme, enabling the extraction of memories from different time steps:

\vspace{-15px}

\begin{align}
\begin{aligned}\label{eq:cell hidden update}
    \operatorname{Cell\, State:}\quad C_t=f_tC_{t-1}+i_tv_tk_t^{\top},\\
    \operatorname{Normalizer\, State:}\quad n_t=f_tn_{t-1}+i_tk_t,\\
    \operatorname{Output\, Gate:}\quad o_t=\sigma(\tilde{o_t}),\quad \tilde{o_t}=W_ox_t+b_o,\\
    \operatorname{Hidden\, State:}\quad h_t=o_t\odot\tilde{h_t},\quad \tilde{h_t}=C_tq_t/ \max \{|n^{\top}_tq_t |,1\},
\end{aligned}
\end{align}
inspired by~\cite{schlag2021linear}, the cell state uses a weighted sum according to proportions, where the forget gate corresponds to the weighted proportion of memory, and the input gate corresponds to the weighted proportion of the key-value pair to satisfy the covariance-based update rule. The mLSTM employs a normalizer that weights key vectors. Ultimately, through normalization and weighted control by the output gate, the hidden state $h_t$ of the network is obtained. The mLSTMs employ matrix values to process memory retrieval, which allows the retrieval process in mLSTMs to be conducted directly through matrix multiplication. The hidden state from timestep t-1 is not included in the processing flow, which greatly enhances the parallelization capability of the mLSTM. The mLSTM introduction of matrix memory and parallelization brings a new level of sophistication to facial expression recognition systems. By employing a matrix memory cell, the mLSTM can store a richer feature representation, capturing the intricate details and variations that define different emotional expressions. Moreover, the parallelization feature of the mLSTM block enables the model to process this complex facial data more efficiently, significantly reduce the computational load.
\subsection{Path Transfer}
By integrating the outcomes from these various views~\cite{schuster1997bidirectional}, a more accurate modeling of the sequence can be achieved. Traditional sequence modeling typically has two path traversal schemes: forward traversal and backward traversal. We have integrated four path scanning schemes: forward and backward bidirectional in the column direction and forward and backward bidirectional in the row direction. The xLSTM incorporates a flip module to achieve a more comprehensive image representation by weighting four paths of the image data.
\subsection{Classification Head}
The output of the xLSTM module will be mapped to the classification dimensions. The current main methods of token aggregation are as follows: 1. Using a learnable [CLS] token placed at the beginning~\cite{dosovitskiy2020image} or middle~\cite{zhu2024vision} of the sequence. 2. Applying average pooling to the entire sequence. 3. Using the average of the first token and the last token as the input for the classification head. In the vast majority of datasets, objects are typically centered around the middle token by default. To avoid this bias and enhance the generality of our model, our experiments adopt the last scheme mentioned. Our loss function is the cross-entropy loss function:

\vspace{-10px}

\begin{align}
\begin{aligned}\label{eq:loss}
    \mathcal{L}=-\sum_{n=1}^{N} y\log(\hat{y})\\
\end{aligned}
\end{align}

\section{Experiments}
\label{sec:experiment}
\subsection{Datasets and Metrics}
We conduct experiments on three  datasets in FER research: CK+~\cite{lucey2010extended}, RAF-DB~\cite{li2017reliable}, and FERplus~\cite{barsoum2016training}. We report the Top-1 accuracy on the seven-category task as the evaluation metric. Here is a brief introduction to the datasets. \textbf{CK+.} The CK+ dataset includes annotations for the following emotions: Anger, Contempt, Disgust, Fear, Happy, Sadness, and Surprise. The CK+ dataset comprises 784 training samples and 197 test samples. \textbf{RAF-DB.} The RAF-DB encompasses seven basic emotional categories: surprise, fear, disgust, happiness, sadness, anger, and neutrality. The training subset encompasses 12,271 images, while the test subset consists of 3,068 images. \textbf{FERplus.} The FERplus dataset is an enhanced version of the original FER dataset. The FERplus dataset categorizes expressions into eight distinct emotions: anger, disgust, fear, happy, sad, surprise, neutral, and contempt. The dataset comprises a total of 28,709 images for training, along with 3,589 images allocated for validation and 3,589 designated for testing purposes.

\subsection{Experiment Settings}
We conduct experiments  with a patch size set to 16x16, the number of stacked xLSTM layers being 26, and the base dimension of the model being 384, which means the dimensions of the query, key, and value vectors are 768. The number of our attention heads is 192.

\subsection{Results}
\label{sec:results}
\begin{table*}[ht!]
\centering\footnotesize
\caption{Results on CK+, RAF-DB, and FERplus. The previous state-of-the-art (SOTA) values are marked with underlines, while the current SOTA values are marked in bold. All reported values are based on the \textbf{``from scratch''} setting.} 
\begin{tabularx}{\linewidth}{c|X|X|X}
\toprule
Method &CK+&RAF-DB&FERplus\\
\midrule
FER-GCN~\cite{fan2018video}&99.54\%&-&-\\
FMPN~\cite{chen2019facial}&98.06\%&-&-\\
FAN~\cite{meng2019frame}&\underline{99.70}\%&-&-\\
SelfCureNet~\cite{wang2020suppressing}&-&\underline{78.31\%}&\underline{83.42\%}\\
ViT~\cite{dosovitskiy2020image}&96.88\%&63.75\%&73.36\%\\
MA-Net~\cite{zhao2021learning}&-&67.48\%&-\\
EAC~\cite{zhang2022learn}&-&73.73\%&75.77\%\\
\midrule
\textbf{xLSTM-FER(ours)}&\textbf{100\%}&\textbf{87.06\%}&\textbf{88.94\%}\\
\textbf{Rank}&\textbf{1}&\textbf{1}&\textbf{1}\\
\bottomrule
\end{tabularx}
\label{tab: Table Results}
\end{table*}

\begin{figure*}[!htb]
    \centering
    \begin{subfigure}{0.30\textwidth}
        \includegraphics[width=\linewidth]{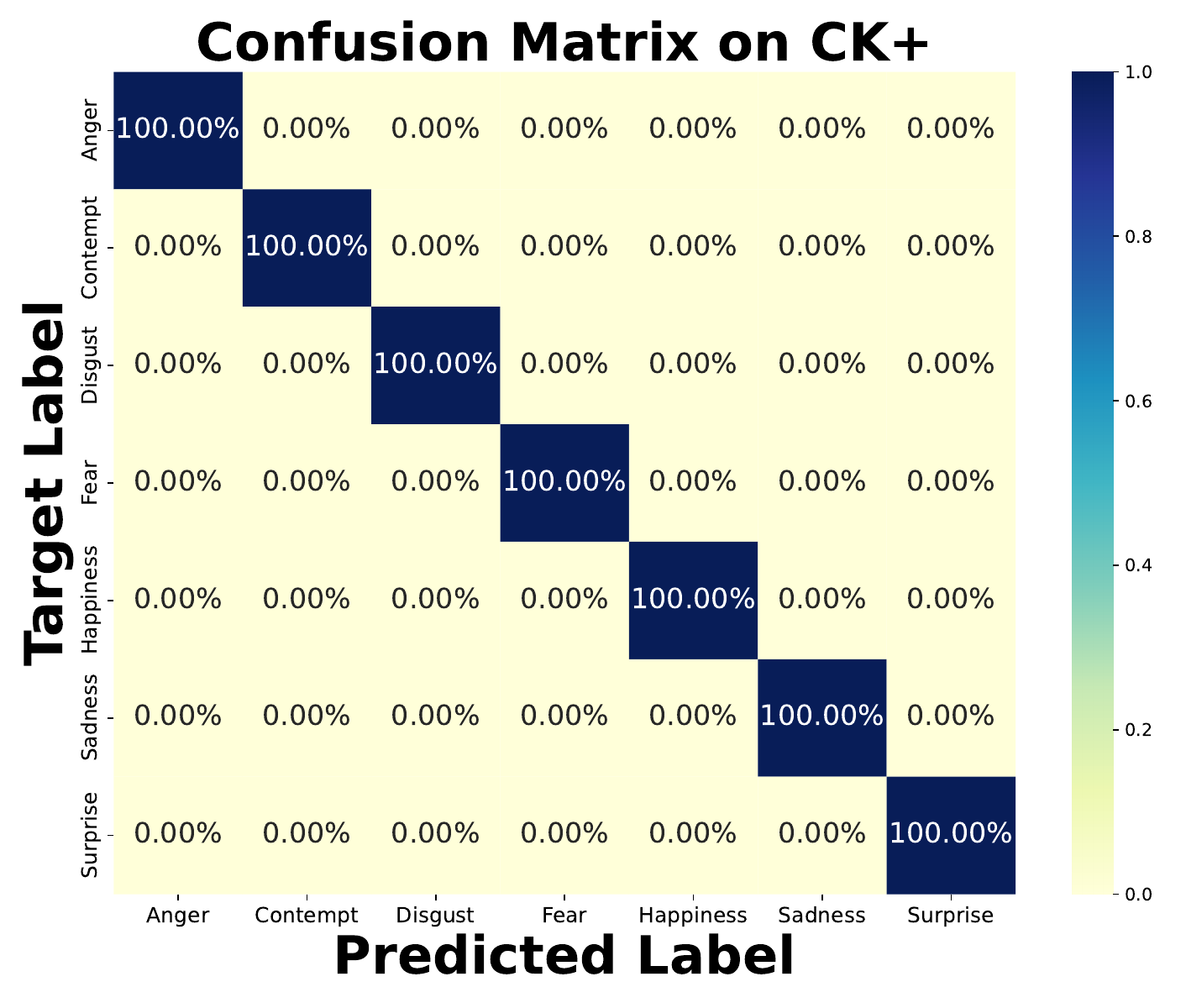}
        \caption{Confusion Matrix on CK+.}
        \label{fig:ck_matrix}
    \end{subfigure}
    \hfill
    \begin{subfigure}{0.30\textwidth}
        \includegraphics[width=\linewidth]{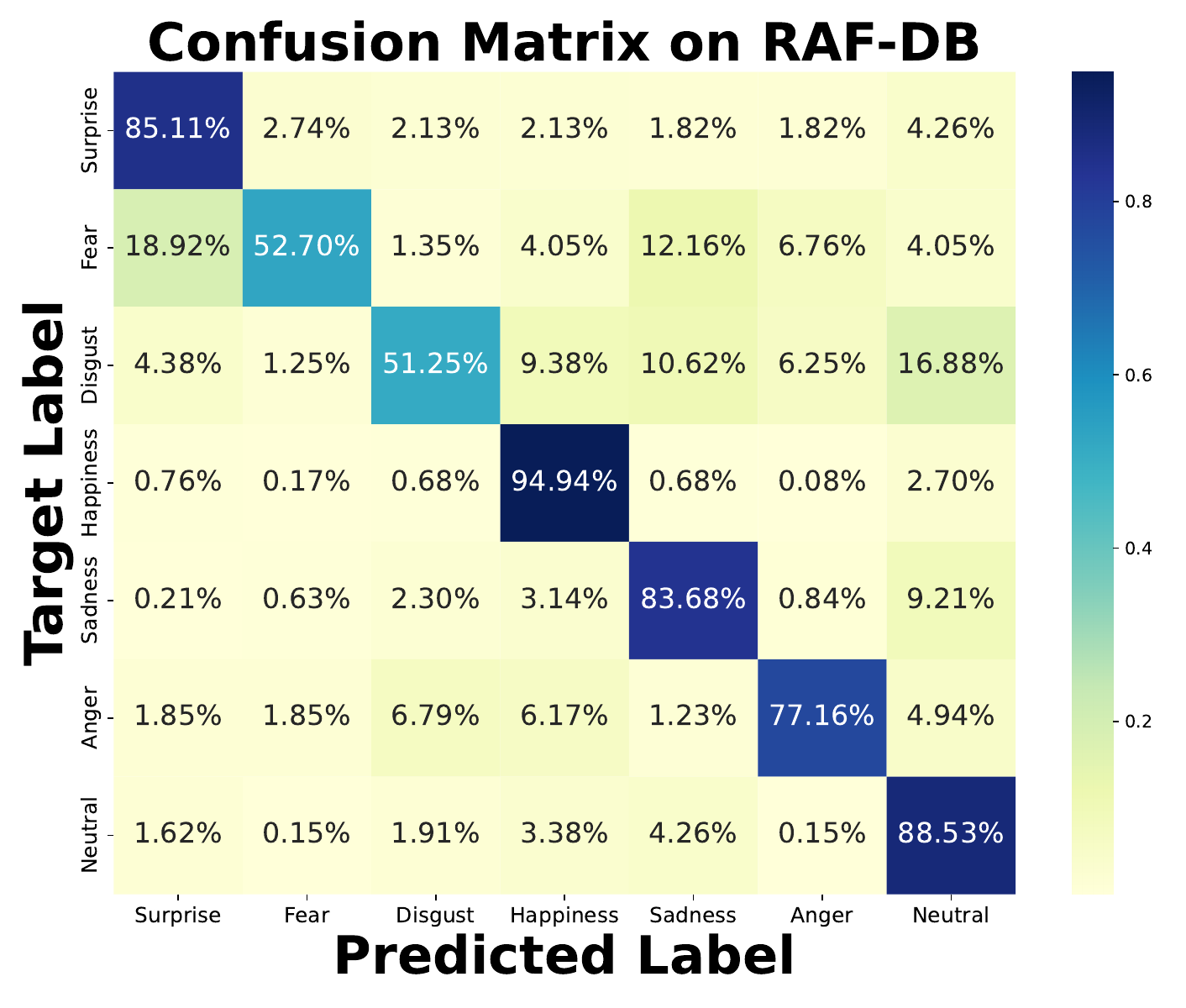}
        \caption{Confusion Matrix on RAF-DB}
        \label{fig:raf_matrix}
    \end{subfigure}
    \hfill
    \begin{subfigure}{0.30\textwidth}
        \includegraphics[width=\linewidth]{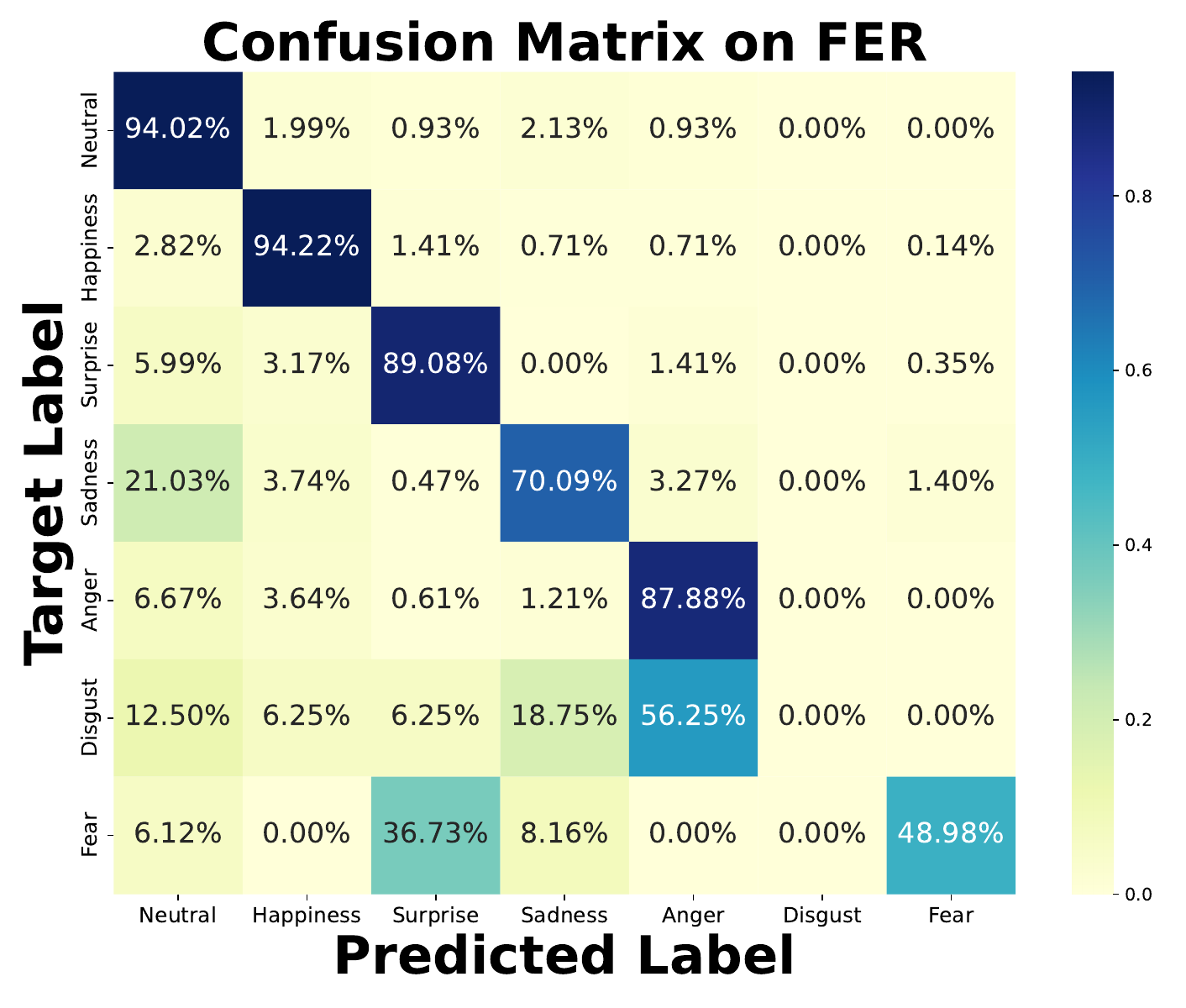}
        \caption{Confusion Matrix on FERplus.}
        \label{fig:fer_matrix}
    \end{subfigure}
    \caption{Confusion Metrics on Datasets.}
    \label{fig:confusion_matrices}
\end{figure*}

We compare xLSTM-FER with the recent CNN-based models such as FER-GCN~\cite{fan2018video}, EAC~\cite{zhang2022learn} and ViT-based face recognition models including ViT~\cite{dosovitskiy2020image}, MA-Net~\cite{zhao2021learning}, and others to verify the effectiveness of xLSTM-FER. All experiments are conducted from scratch. The experimental results are shown in Table~\ref{tab: Table Results}.

\textbf{Results on CK+.} In the CK+ dataset, the outcomes presented in Table 7 reveal that our technique, xLSTM-FER, has pioneered a perfect accuracy rate of 100\% for classifying facial expressions. The confusion matrix in Fig.~\ref{fig:ck_matrix} indicates that xLSTM-FER has achieved 100\% accuracy across all categories. These outcomes excel over the sota methods in the realms of both video and image-based facial expression analysis. Because xLSTM-FER successfully captures the interdependencies among different patch blocks.

\textbf{Results on RAF-DB.} xLSTM-FER achieves an impressive overall accuracy of 87.06\% and shows a 14\% improvement over the previous sota values, demonstrating superior performance compared to other models. In contrast, ViT only achieves a lower accuracy of 63.75\%, and another visual transformer-based FER model, MA-Net, does not perform well on this in-the-wild dataset. xLSTM-FER achieves a competitive accuracy of 87.06\%, indicating its robust performance compared to current state-of-the-art methods.

\textbf{Results on FERplus.} On the FERplus dataset, our model has outperformed all contemporary methods, attaining an accuracy rate of 88.94\%. xLSTM-FER shows a 4.5\% improvement compared with previous sota values. This confirms that the synergistic effect of the memory gating and attention mechanisms within xLSTM-FER can achieve an accurate representation of facial images.

\subsection{Case Analysis}
\vspace{-20px}
\begin{figure*}[!htb]
    \centering
    \includegraphics[width=0.8\textwidth]{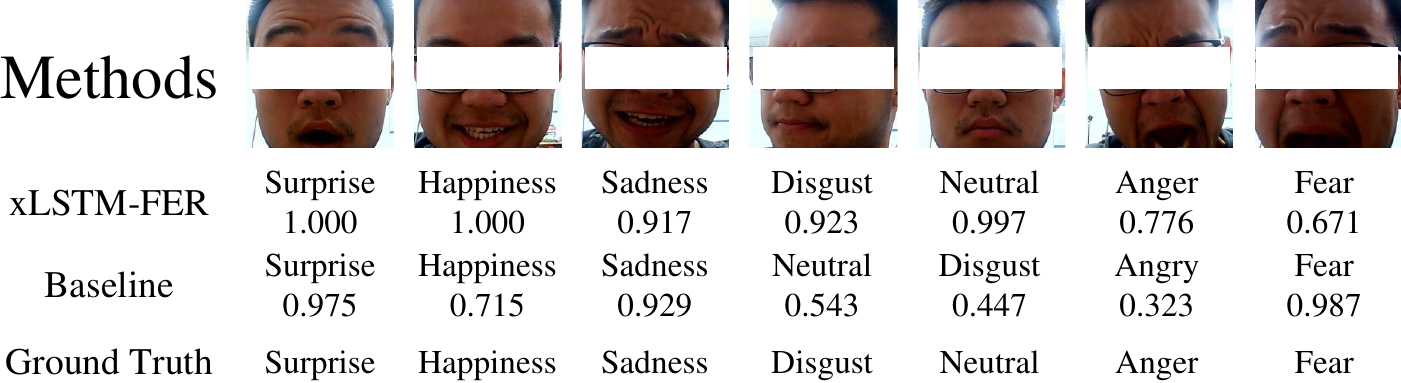}
    \caption{A case study on the accuracy of xLSTM-FER compared to the baseline model (EAC~\cite{zhang2022learn}) in real-world examples.\label{fig:case}
     }
\end{figure*}
\vspace{-20px}

To further verify the advantages of xLSTM-FER over the baseline, we test several photos in the learning environment. The test results are shown in Fig.~\ref{fig:case}. We find that, except for the ``Fear'' and ``Sadness'' categories, xLSTM-FER can provide more accurate predictions with higher confidence compared to EAC in other categories. This indicates that the memory network of xLSTM in its image extraction approach can adapt to the real-world needs of students' FER tasks even with the linear complexity.

\section{Conclusion}
\label{sec:conclusion}
To overcome the quadratic complexity in traditional student facial expression recognition, this paper presents xLSTM-FER, which has profound implications for the assessment of learning experiences and emotional states. The innovative approach of xLSTM-FER in segmenting input images into patches and processing them through a stack of xLSTM blocks. Our experimental results on CK+, RAF-DB, and FERplus not only validate the potential of xLSTM-FER in student facial expression recognition tasks but also highlight its competitive performance when compared to state-of-the-art methods on standard datasets. The linear computational and memory complexity of xLSTM-FER is a significant advantage, making it exceptionally well-suited for processing high-resolution images, which is essential for the clear and detailed capture of student expressions. We are confident that with further optimization and fine-tuning, xLSTM-FER will evolve as a significant tool in student expression recognition.

\section*{Acknowledgement}
The research project is supported by the National Natural Science Foundation of China (No. 62207028), partially by Zhejiang Provincial Natural Science Foundation (No. LY23F020009), and the Key R\&D Program of Zhejiang Province (No. 2022C03106), and Scientific Research Fund of Zhejiang Provincial Education Department (No. 2023SCG367).
\bibliographystyle{splncs04}

\end{document}